\documentclass[10pt,twocolumn,letterpaper]{article}

\usepackage[pagenumbers]{cvpr} % To force page numbers, e.g. for an arXiv version

\usepackage[accsupp]{axessibility}  % Improves PDF readability for those with disabilities.
\usepackage{graphicx}
\usepackage{amsmath}
\usepackage{amssymb}
\usepackage{booktabs,cancel}
\usepackage{multirow}
\usepackage{multicol}
\usepackage{array}
\usepackage{algorithm}
\usepackage{algorithmic}
\usepackage{listings}
\usepackage{xcolor}
\usepackage{setspace}
\definecolor{codeblue}{rgb}{0.25,0.5,0.25}
\usepackage[pagebackref,breaklinks,colorlinks]{hyperref}

\usepackage[capitalize]{cleveref}
\crefname{section}{Sec.}{Secs.}
\Crefname{section}{Section}{Sections}
\Crefname{table}{Table}{Tables}
\crefname{table}{Tab.}{Tabs.}

\def\cvprPaperID{7194} % *** Enter the CVPR Paper ID here
\def\confName{CVPR}
\def\confYear{2022}

\begin{document}

\title{Style Transformer for Image Inversion and Editing}

\author{$\text{Xueqi Hu}^1$, $\text{Qiusheng Huang}^1$, $\text{Zhengyi Shi}^1$, $\text{Siyuan Li}^1$, $\text{Changxin Gao}^3$, $\text{Li Sun}^{1,2}$\footnotemark[1], $\text{Qingli Li}^1$\\
$^1$Shanghai Key Laboratory of Multidimensional Information Processing,\\
$^2$Key Laboratory of Advanced Theory and Application in Statistics and Data Science,\\ East China Normal University, Shanghai, China\\
$^3$Huazhong University of Science and Technology, Wuhan, China}


\twocolumn[{
\renewcommand\twocolumn[1][]{#1}
\maketitle
\vspace{-0.5 cm}
\begin{center}
    \centering
    \includegraphics[width=1\textwidth]{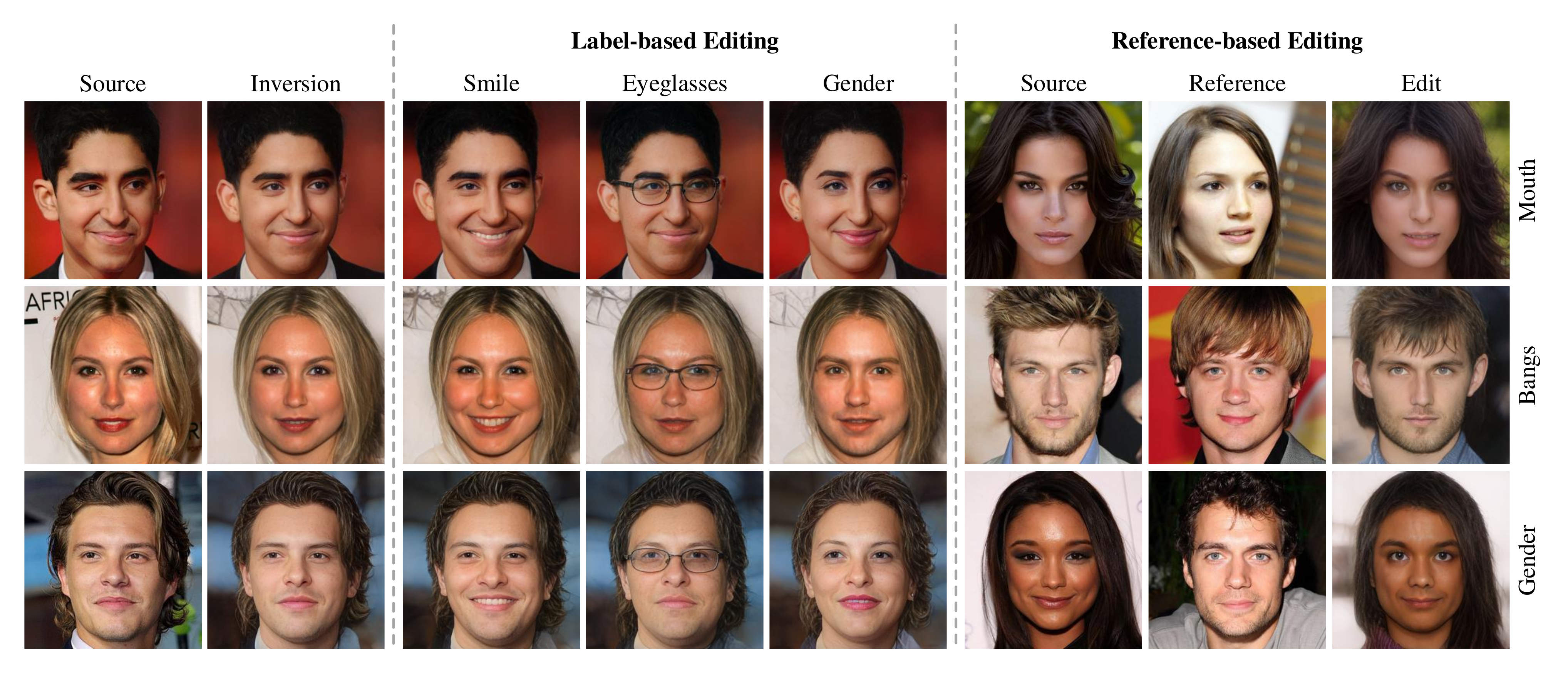}
    \captionof{figure}{Image inversion and editing results of our model on CelebA-HQ dataset. From left to right, we show the inversion, label-based editing and reference-based editing. %from left to right. 
    For the reference-based editing, the three columns are the source, reference and edit images, and each edit image takes the style of the reference while maintaining the source content.} 
    \label{fig:teaser}
\end{center}
\vspace{0.2cm}
}]

\maketitle

%%%%%%%%% ABSTRACT
\begin{abstract}
   Existing GAN inversion methods fail to provide latent codes for reliable reconstruction and flexible editing simultaneously. This paper presents a transformer-based image inversion and editing model for pretrained StyleGAN which is not only with less distortions, but also of high quality and flexibility for editing. The proposed model employs a CNN encoder to provide multi-scale image features as keys and values. Meanwhile it regards the style code to be determined for different layers of the generator as queries. It first initializes query tokens as learnable parameters and maps them into $W^+$ space. Then the multi-stage alternate self- and cross-attention are utilized, updating queries with the purpose of inverting the input by the generator. Moreover, based on the inverted code, we investigate the reference- and label-based attribute editing through a pretrained latent classifier, and achieve flexible image-to-image translation with high quality results. Extensive experiments are carried out, showing better performances on both inversion and editing tasks within StyleGAN. Codes are available at \href{https://github.com/sapphire497/style-transformer}{https://github.com/sapphire497/style-transformer}.
   \noindent\footnotetext{Corresponding author, email: sunli@ee.ecnu.edu.cn. This work is supported by the Science and Technology Commission of Shanghai Municipality (No.19511120800) and Natural Science Foundation of China (No.61302125).}
\end{abstract}

%%%%%%%%% BODY TEXT
\section{Introduction}
Generative Adversarial Network (GAN) \cite{brock2018large,oord2017neural,razavi2019generating} has been significantly improved during recent years. Particularly, with the help of AdaIN \cite{huang2017arbitrary} or it variation ModulatedConv, StyleGAN \cite{karras2019style,karras2020analyzing} is able to synthesize high resolution images with moderate quality. Therefore, utilizing the pretrained and fixed StyleGAN for downstream tasks becomes a hot research topic, especially in the editing task of image-to-image (I2I) translation \cite{harkonen2020ganspace,shen2020interfacegan,shen2021closed,wu2021stylespace,yao2021latent,abdal2021styleflow}. 

To edit a given real-world image, we first need to find out its input noise vector $z$ or intermediate latent code $w$, which can faithfully reconstruct the specified real image using the pretrained generator. Then, the code is modified by an offset corresponding to the target attribute, so that it can be mapped into an edited image, while preserving the original details. Despite of the great efforts, inverting \cite{abdal2019image2stylegan,abdal2020image2stylegan++,zhu2020domain,richardson2021encoding,tov2021designing} or editing \cite{harkonen2020ganspace,shen2020interfacegan,wu2021stylespace,shen2021closed,abdal2021styleflow} images for StyleGAN is still challenging due to following reasons. 
First, there are several candidate latent embeddings. Existing methods \cite{tov2021designing,zhu2020improved,wu2021stylespace} reveal that different choices on them are critical. Compared to $Z$ or $W$ space with a single 512-d vector, $W^+$ has the enough ability to describe image details, therefore it is suitable for inversion. In $W^+$, each image is represented by 18 different codes, and each of them is 512-d. They are given to the generator to formulate features and final synthesis from low to high resolutions in sequence. However, the code in $W^+$ can not be well edited unless imposing enough regularization. 
Second, the distribution in $W$ or $W^+$ are highly complex. Real images only lie on the manifold in the space \cite{tov2021designing}. Moreover, different dimensions are often entangled for a single attribute, making independent editing difficult.

This paper aims to improve the encoder-based image inversion and editing for StyleGAN at the same time. Inspired by the great success of transformer in image classification \cite{dosovitskiy2020image,liu2021swin} and object detection \cite{carion2020end,zhu2020deformable}, we utilize it to find the appropriate latent code in $W^+$ space for image inversion and editing tasks. The basic idea is to regard latent codes in different generation stages as query tokens, and image features at different spatial positions as keys and values, then perform the multi-head cross-attention to update the queries in an iterative way. Meanwhile, before the cross-attention, the queries are also allowed to access others through the self-attention, to enhance the regularization on them, so the final codes given to the generator become tightly linked. 

Particularly, queries first interact with image features (keys) by comparing similarities between each query-key pair. Then they are organized into the attention matrix to dynamically weight the features (values) and update queries for the transformer block in next stage. The image features, used as keys and values, are obtained by a CNN encoder. To capture the image details at different resolutions, we employ a two-pyramid encoder proposed in \cite{richardson2021encoding} to provide multi-scale features as keys and values. Note that our model has the multiple cross-attentions from low to high resolutions, and the style queries are gradually updated by features at different scales. Therefore, general contents in queries are first formulated, and then refined by details in the higher resolution. After several times self- and cross-attentions, queries absorb enough details from the input image, so they can be utilized to invert it by the pretrained generator. 

We are further interested in the way to edit the codes for translating a specified attribute. Traditional approaches \cite{harkonen2020ganspace,shen2020interfacegan,wu2021stylespace} assume the linear separations in the latent space for a binary attribute, so inverted code from different images are edited by the same direction. We argue that the identical direction is not optimal for the editing quality, and may reduce the result diversity. Inspired by \cite{choi2020stargan,huang2021bridging}, we divide the image editing in StyleGAN into two different types. One is label-based editing, in which only target label is specified. The other is reference-based editing, which requires another image to supply the desired style. For the former, a pretrained non-linear latent classifier is used to determine the direction. it computes a loss for the inverted code according to the target label, and its gradient is back-propagated to the code, giving the editing direction. In the latter case, we want to determine the exact editing vector from the reference. Therefore, the inverted code from the source is used as query, and from the reference as key and value. The cross-attention is performed between them. The parameters in the attention module is trained under the supervision from the latent classifier, encouraging the edited attribute to be similar with the reference and other attributes without any changes. The proposed editing method is able to give the diverse results while maintaining the quality of image.

The contribution of the paper is summarized into following aspects. First, we propose novel multi-stage style transformer in $W^+$ space to invert image accurately. The transformer includes the self- and cross-attention modules, in which the style queries gradually get updated from the multi-scale image features. Second, we characterize the image editing in StyleGAN into label-based and reference-based, and use a non-linear classifier to generate the editing vector. Diverse and fidelity editing results are obtained.

\section{Related Works}
\textbf{GAN inversion} is first proposed in \cite{zhu2016generative} and becomes important due to the wide applications of some recent generators. There are basically two ways, either encoder-free or encoder-based. The former does not have any training parameters, and the latent codes are directly optimized by the gradient mainly from reconstruction loss. To deal with the complex latent structure, Abdal \etal \cite{abdal2019image2stylegan} invert a real image in $W^+$ space, and use pixel-wise MSE and perceptual loss with Adam \cite{kingma2014adam} to tune the code. Image2StyleGAN++ \cite{abdal2020image2stylegan++} extends the code space to the layerwise additive noise vector to decrease the distortion. Although such a method can reliably find the code through multi-step iterations, it is inefficient and its code lacks the editability. 

In contrast, the encoder-based method intends to train a common model to achieve inversion for all images. It improves the editing ability and is efficient during inference. IDInvert \cite{zhu2020domain} utilizes a CNN as encoder to output the code. Except the reconstruction and perceptual loss, it is trained by an extra adversarial loss. pSp \cite{richardson2021encoding} designs a two-pyramid encoder to provide multi-scale features, and maps them to the style vector through multiple convolution layers. Benefiting from strong features, pSp achieves less distortion. ReStyle \cite{alaluf2021restyle} uses the encoder to give the residual style to refine the inversion in the iterative way. E4e \cite{tov2021designing} analyzes the distortion-editability trade off for inversion and editing tasks in $W^+$. It sacrifices the inversion accuracy to improve the editablity, constraining the codes for different layers close with each other. Kim \etal \cite{kim2021exploiting} and Wang \etal \cite{wang2021HFGI} depart from $W^+$ space and enhance the code with spatial dimensions, so that more information are given to the generator to lower down the distortions. Compared to previous works, our method lies strictly in $W^+$, and it is able to achieve minimal distortion and high quality editing at the same time.

\textbf{Latent code manipulation} for pretrained StyleGAN is often used to edit the attribute and achieve I2I translation, either in the supervised or unsupervised manner. GANSpace \cite{harkonen2020ganspace} and Sefa \cite{shen2021closed} adopt PCA to find the principal directions in $W$ space. They are responsible for controlling the pose, gender or background. Note that for a particular attribute, these works specify the same direction on all latent codes to realize editing. Voynov and Babenko \cite{voynov2020unsupervised} train a simple module to edit the input, and use a reconstructor in pixel domain to interpret the editing, finding noticeable directions. LatentCLR \cite{yuksel2021latentclr} builds a learnable direction model to edit the code and uses contrastive loss to train it. Hence, these two models give different images unique editing directions. All the above works are unsupervised, without requiring attribute label for editing. But only limited directions for some attributes can be found. 

The supervised methods can identify directions for more attributes, especially the local ones. InterfaceGAN \cite{shen2020interfacegan} trains the linear binary SVM in latent space to obtain a separation plane, whose normal vector controls its corresponding attribute. StyleSpace \cite{wu2021stylespace} finds the control direction in a precise way guided by the semantic mask. Moreover, they propose to edit the code in $S$ space which is defined by the affine layer after $W$. Wang \etal \cite{wang2021attribute} further extend $S$ space by tracing back the gradient flow to its previous stage, making the change more accurately. Note that these works still share the same editing direction for all images. Recently, StyleFlow \cite{abdal2021styleflow} conditionally manipulates the images using Continuous Normalizing Flows (CNF). Yao \etal \cite{yao2021latent} propose a latent transformation module to generate adaptive directions for different images. Wang \etal \cite{wang2021HFGI} utilize a CNN encoder to provide multi-scale features to supplement the $1\times1$ style vector, which actually adapts directions at different locations. 

However, previous works only deal with the label-based editing. Collins \etal \cite{collins2020editing} apply k-means clustering on features to obtain channel-wise masks, determining which channels are locally semantic-aware. The cluster memberships of the reference further guide local attribute editing for the source. Different from above works, our work is strictly in $W^+$, and we realize both the label- and the reference-based editing.

\begin{figure*}[ht]
\centering
\includegraphics[width=.9\textwidth]{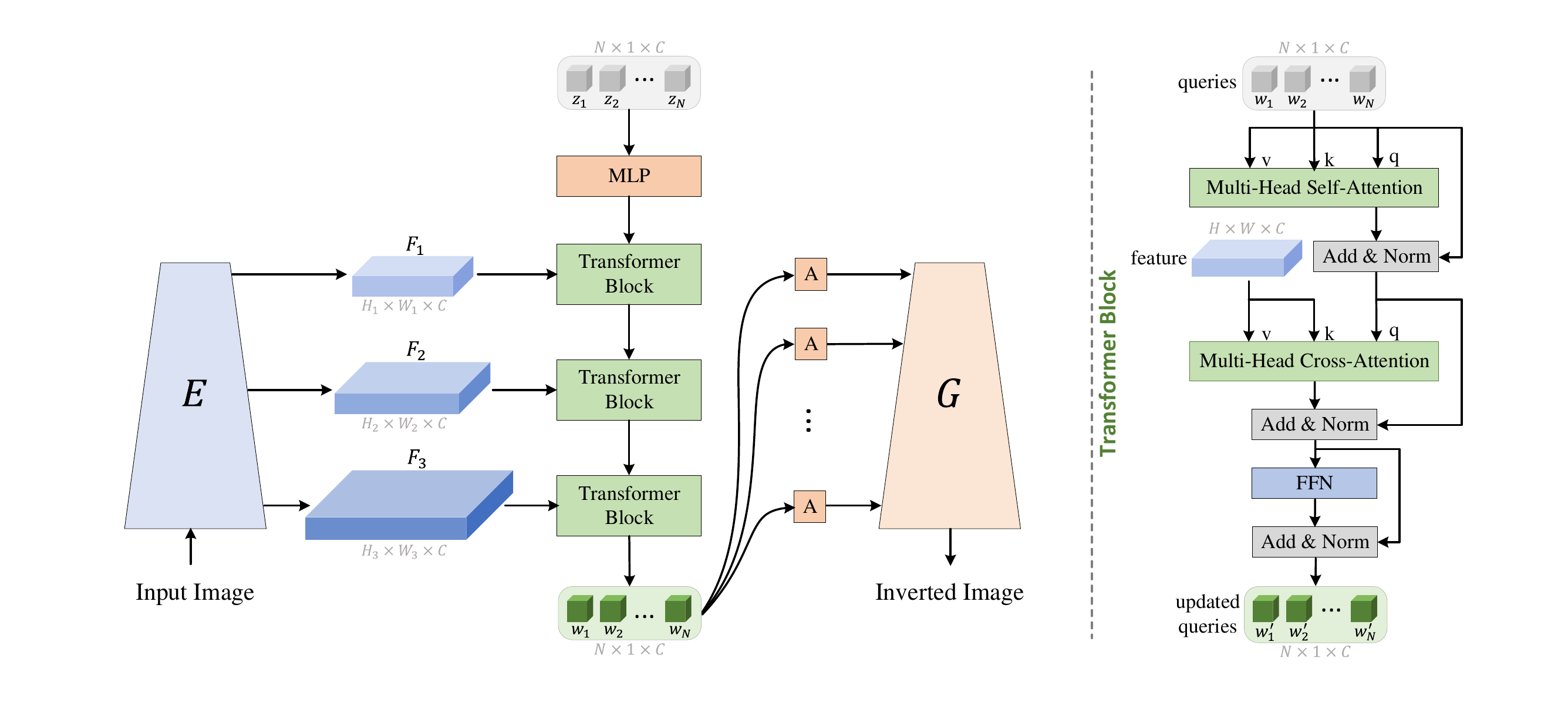}
\caption{The overall framework of Style Transformer for image inversion. We build the multi-stage transformer-based model to update the code in $W^+$ space. Details within the transformer block are depicted on the right. Each has a multi-head self- and cross-attention block, following the common routine in transformer model.}
\label{fig:fig1}
\vspace{-0.3cm}
\end{figure*}

\section{Framework of Style Transformer}
%\subsection{Preliminaries on StyleGAN and Transformers}
We aim to achieve accurate image inversion for StyleGAN by our proposed style transformer in $W^+$ space. Given a real image $I\in\mathbb{R}^{H\times W\times 3 }$, our model is able to specify $N$ different style vectors denoted by $w_n\in\mathbb{R}^{512}$, where $n=1,2,\cdots, N$ is the index of the vector injected into the different stages of the generator $G$. For simplicity, we use $w\in\mathbb{R}^{N\times512}$ without any index to represent all $w_n$. Note that in StyleGAN2, $w_n$ is first projected by the affine layer $A$, then it affects the corresponding layers by modulating on the convolution kernels.

\cref{fig:fig1} illustrates the overview of the proposed framework. The input image $I$ is encoded by $E$, generating a series of image features $F_1$ to $F_3$ in multi-resolutions \cite{richardson2021encoding}. $N$ different queries, output from an MLP, access these features through the transformer blocks in a sequential way, forming the final code $w$ for the generator. The initial input $z_n\in \mathbb{R}^{512}$ to the MLP is also learnable, and it is gradually updated into $w$, which is suitable for inverting $I$. By training all parameters including the transformer blocks, the encoder $E$, the MLP and the initial $z_n$, the pretrained $G$ can utilize the final $w$ to reconstruct input $I$ with minimal distortions. 

\subsection{Style Transformer Block}
The style transformer block is the key component for image inversion. The same structure is applied for 3 times in the model, exploiting the image details from $F_1$ to $F_3$, respectively. The specific design within the block is shown on the right of \cref{fig:fig1}. Basically, there are two types of attention, which are multi-head self-attention and cross-attention. In addition, we follow the design routines for transformer, incorporating the residual connection, the normalization and FFN module into the block.

\textbf{Style query initialization.} Given a single style code $w$, high fidelity image can be synthesized by StyleGAN generator. However, $W^+$ space needs $N$ different style vectors to reconstruct one image, and they essentially describe the details at different scales, therefore are employed to affect features of different resolutions in the generator. A common choice in transformer decoder is to randomly initialize beginning query tokens and keep them as learnable parameters in the model. However, considering the fact that code distribution in $W$ space is complex and far from Gaussian prior, we utilize the pretrained MLP in StyleGAN to first map each individual code $z_n$ to the beginning style query $w_n$ in $W$ space, and then update $w_n$ through the self and cross-attention operations. Note that $z_n\sim N(0, I)$ is sampled from standard Gaussian, and set as model parameters. Moreover, the pretrained MLP is finetuned during training. 

\textbf{Multi-Head Self-Attention. }The self-attention is performed among $N$ different query tokens $q_1,q_2,\cdots,q_N$. It intends to find the potential relation between any pair of them, and route the value to connect them. We denote all of $q$ as $X_q\in \mathbb{R}^{N\times 512}$. The query $Q$, key $K$ and value $V$ are all projected from $X_q$ according to \cref{eq:eq1}. Note that $W_Q^{self}$, $W_K^{self}$ and $W_V^{self}\in\mathbb{R}^{512\times 512}$ are learnable projection heads in the self-attention module, which do not change the feature dimension.
\begin{equation}\label{eq:eq1}
  Q=X_q W_Q^{self},\quad K=X_q W_K^{self}, \quad V=X_q W_V^{self}
\end{equation}
The multi-head attention operation is formulated as in \cref{eq:eq2}, where $Q_i$, $K_i$ and $V_i$ are query, key and value in the $i$th head, and $Attn$ is result from that head. The feature dimension $d=512/H$, and $H$ is the number of attention heads.
\begin{equation}\label{eq:eq2}
  Attn(Q,K,V)= \text{Softmax}(\frac{Q_i K_i^T}{\sqrt{d}}) V_i
\end{equation}
The final update on $X_q$ from the self-attention is $MHA$ in \cref{eq:eq3}. $W^o\in\mathbb{R}^{512\times 512}$ is also learnable, being responsible for fusion the results $Attn$ from different heads.
\begin{equation}\label{eq:eq3}
  MHA(Q,K,V)= [Attn(Q_i, K_i, V_i)]_{h=1:H} W^o 
\end{equation}

\textbf{Multi-Head Cross-Attention.} The self-attention does not involve any image feature in its computation. Therefore, we further design the cross-attention for the inversion task, so that the query tokens can obtain information from image features $F_1$, $F_2$ and $F_3$ in different resolutions. In the cross-attention, features of key and value are from the encoder $E$, while queries are computed by the linear projection on the previous results from self-attention block. Particularly, we have the query, key and value according to \cref{eq:eq4}, where $W_Q^{crs}$, $W_K^{crs}$ and $W_V^{crs}$ share the similar settings with the self-attention.
\begin{equation}\label{eq:eq4}
  Q=X_q W_Q^{crs},\quad K=F_i W_K^{crs}, \quad V=F_i W_V^{crs}
\end{equation}
The multi-head cross-attention is carried out in the same way as is shown in \cref{eq:eq2} and \cref{eq:eq3}. After that, the updated query tokens are given to an FFN to refine itself, and the results are further passed to the transformer block in the next stage, mining the details from finer resolution features.

\subsection{Training Objectives for Image Inversion}
During training, the backbone $G$ of StyleGAN (including the affine layer $A$) is strictly fixed. The gradients from the loss only tune other parameters. Note that we use the same training objectives as pSp \cite{richardson2021encoding}. Particularly, to give the accurate reconstruction, the $L_2$ loss between the input $I$ and its inverted version $\hat{I}$ from $G$ is calculated. Meanwhile, LPIPS \cite{zhang2018unreasonable}, a similarity metric, which is computed based on the features in an Inception net $F(\cdot)$, is also adopted, specifying another objective $L_{LPIPS}=\Vert F(I)-F(\hat{I}) \Vert_2$. Additionally, to keep the identity of the inverted image, we incorporate a pretrained ArcFace model \cite{deng2019arcface} $R(\cdot)$ for the ID loss $L_{ID}=1-\langle R(I),R(\hat{I}) \rangle$, so that the cosine similarity of $I$ and $\hat{I}$ can be maximized. Notice that we do not adopt any adversarial loss during training.

\begin{figure}[t]
\centering
\includegraphics[width=.8\columnwidth]{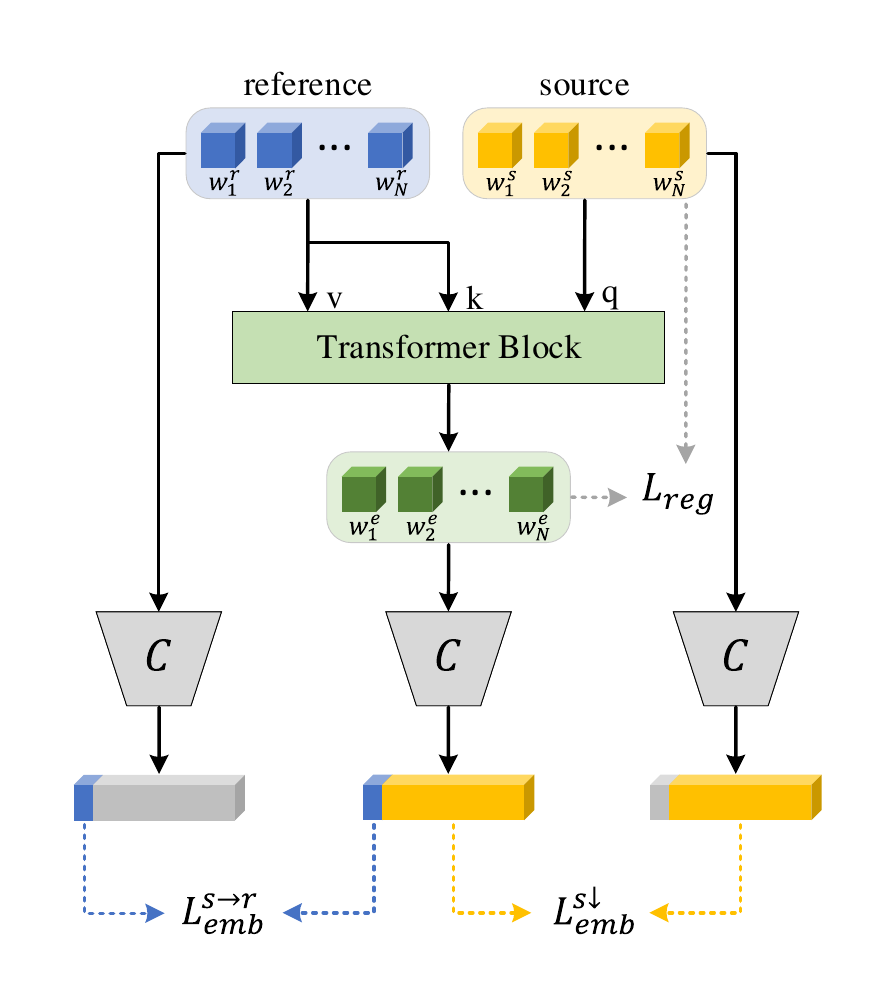}
\vspace{-0.1cm}
\caption{Reference-based editing module and its training strategy. The inverted codes $w^s$ and $w^r$, from the source and reference, are given to the transformer module $T$, specifying the code $w^e$ for edited image. $C$ is an attribute classifier in $W^+$, by which we constrain the editing attribute being similar with the reference, while others staying the same as the source. }
\label{fig:fig2}
\vspace{-0.2cm}
\end{figure}

\section{Image Editing in Style Transformer}
Image editing by the fixed StyleGAN is an important application not only for itself, but also for evaluating the quality of image inversion. The low distortion is only the one aspect, the flexible and high fidelity editing are also important. As is described in \cite{choi2020stargan, huang2021bridging}, there are two types of editing, either through the target label or a reference image in the desired domain. Previous works \cite{shen2020interfacegan, wu2021stylespace, yao2021latent} focus on the former, but few works deal with the referenced-based editing, which potentially provides diverse results. 

Typically, given the inverted style code $w^s\in\mathbb{R}^{N\times512}$ for the source $I_s$, and a desired target attribute, we need to determine an offset $\Delta w$, so that $w^e =w^s+\Delta w$ can be mapped to an edited image $\tilde{I}=G(w^e)$ with the desired attribute different from $I_s$, but keep content of $I_s$. In the reference-based editing, another image $I_r$ is given as an extra input. Since our style transformer can invert image with negligible distortion, we train a latent classifier $C$ for $K$ binary attributes in $W^+$ space to guide the editing like \cite{yao2021latent}. Concretely, given a code $w$ inverted from an image, the classifier computes several embedding features $C_f^k$ corresponding to the $k$th attributes, and the final logits $C_l^k$ for the BCE loss $L_{bce}$. During editing, $C$ is fixed to evaluate $w^e$. 

\subsection{Reference-based Editing}\label{sec:refbased}
\textbf{Module design.} We design a simple module $T$ to translate a particular attribute according to the inverted code $w^r$ from reference $I_r$. Since both $w^s$ and $w^r$ represent images with almost no distortions, these codes contain enough information about the edited attribute. $T$ should be able to specify $\Delta w$ based on $w^r$ and $w^s$. Again, a cross-attention structure is chosen, as is shown in \cref{fig:fig2}. $Q=w^sW_Q^{edt}$ are used as a series of query tokens, while $K=w^rW_K^{edt}$ and $V=w^rW_V^{edt}$ are key and value tokens, projected from $w^r$. According to \cite{wu2021stylespace,shen2020interfacegan}, some local attributes are only depended on a single $w_n$ for a particular resolution in $G$. So we choose a routing scheme in \cite{locatello2020object} different from \cref{eq:eq2}. The idea is to make $\text{Softmax}_Q$ normalize over queries not keys. Then re-norm the matrix over keys by $\text{Norm}_K$ as is \cref{eq:eq5}. This strategy assigns the value feature $V$ to queries in the unique way, so that a value token from $w^r$ affects only a few tokens in $w^s$. 
\begin{equation}\label{eq:eq5}
  %Attn(Q,K,V) 
   T(w^s,w^r)=\Delta w= \text{Norm}_K(\text{Softmax}_Q(\frac{Q K^T}{\sqrt{d}})) V
\end{equation}

\textbf{Loss designs.} To guarantee the $k$th attribute editing results, we design the following loss terms to train the projection head in $T$. Particularly, we constrain the code $w^e$ after editing by $L_{emb}^{s\rightarrow r}$ as is shown in \cref{eq:eq6}:
\begin{equation}\label{eq:eq6}
  L_{emb}^{s\rightarrow r}= \Vert C_f^k(w^e)-C_f^k(w^r) \Vert_2
\end{equation}
Here $C_f^k$ is the $k$th attribute embedding from the pretrained latent classifier $C$. $L_{emb}^{s\rightarrow r}$ ensures the edited attribute to be similar with $I^r$. At the same time, other attributes denoted by $\cancel{k}$ should stay close with the source $I_s$, giving $L_{emb}^{s\downarrow}$ in \cref{eq:eq7}: 
\begin{equation}\label{eq:eq7}
  L_{emb}^{s \downarrow}= \Vert C_f^{\cancel{k}}(w^e)-C_f^{\cancel{k}}(w^s) \Vert_2
\end{equation}
Finally, we regularize $L_{reg}=\Vert \Delta w \Vert_2=\Vert w^e-w^s \Vert_2$, so edited image $\tilde{I}$ does not change much.

\subsection{Label-based Editing}\label{sec:labelbased}
Compared to reference-based, label-based editing is relatively easy. So we adopt an encoder-free method to edit $w$ based on the latent classifier $C$. We emphasize that for each $I_s$, there should be a unique direction $n_{\Delta w}^k$ for the $k$th attribute editing, which is determined by the gradient back-propagated from the classifier $C$. Note that the first-order gradient on $w$ is $g=\nabla_w L_{bce}(C_l^k(w^s),y_t)$, and the direction becomes $n_{\Delta w}^k=-g/||g||_2$. Here $y_t$ is the target label, and $C_l^k(w^s)$ is the logits after sigmoid. 

We also investigate the method based on the second-order derivative $H$, which is the Hessian matrix. Similar with \cite{miyato2018virtual}, a randomly sampled unit vector $d$ is first obtained, then it is scaled by a small number $\xi$. Then we evaluate the Hessian vector product by \cref{eq:eq9}. According to power iteration, $d\leftarrow Hd$ converges to the dominant eigenvector, so we let $g=Hd$.
\begin{equation}\label{eq:eq9}
\resizebox{1.0\hsize}{!}{$Hd\approx\frac{\nabla_r L_{bce}(C^k(w^s+r),y_t)|_{r=\xi d}-\nabla_r L_{bce}(C^k(w^s),y_t)|_{r=0}}{\xi}$}
\end{equation}

\begin{figure*}[!htbp]
\centering
\includegraphics[width=1\textwidth]{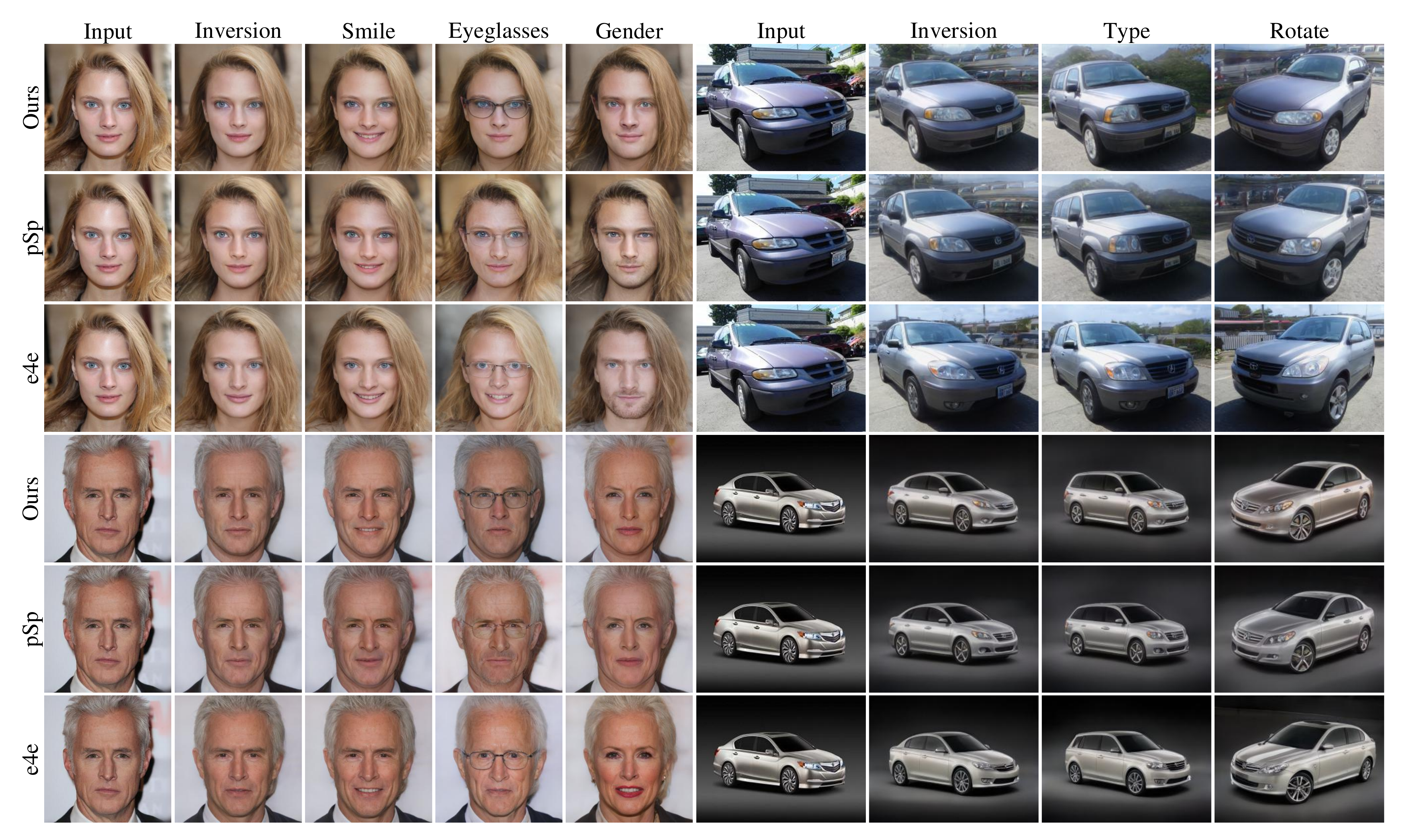}
\vspace{-0.1cm}
\caption{Qualitative results of image inversion. Our method is compared with pSp and e4e. Besides the inverted image, we also list images edited by InterFaceGAN \cite{shen2020interfacegan} for faces. For cars, we use the directions provided in GANSpace \cite{harkonen2020ganspace} for editing.}
\label{fig:inversion}
\vspace{-0.2cm}
\end{figure*}

\begin{table*}[t]
\renewcommand\arraystretch{1.1}
\centering
\resizebox{1\textwidth}{!}{
% \begin{tabular}{p{9mm}<{\centering}p{10mm}<{\centering}|p{7mm}<{\centering}p{8mm}<{\centering}p{6mm}<{\centering}p{7.5mm}<{\centering}|p{6mm}<{\centering}p{7mm}<{\centering}}
\begin{tabular}{cc|cccc|cc|ccc}
\hline
\multirow{2}{*}{\textbf{Domain}} & \multirow{2}{*}{\textbf{Method}} & \multicolumn{4}{c|}{\textbf{Inversion}} & \multicolumn{2}{c|}{\textbf{Editing}} & \multicolumn{3}{c}{\textbf{Model Size}} \\ \cline{3-11} 
                      &      & MSE$\downarrow$  & LPIPS$\downarrow$ & FID$\downarrow$ & SWD$\downarrow$ & FID$\downarrow$ & SWD$\downarrow$ & Params(M)$\downarrow$ & FLOPs(G)$\downarrow$ & Time(s)$\downarrow$     \\ \hline
\multirow{3}{*}{Face} & pSp  & 0.037          & 0.169          & 31.52          & 15.07          & 46.64          & 29.05          & 267.3         & 72.55          & 0.0668\\
                      & e4e  & 0.050          & 0.209          & 36.16          & 17.25          & 47.45          & 25.10          & 267.3         & 72.55          & 0.0659\\
                      & Ours & \textbf{0.036} & \textbf{0.166} & \textbf{28.31} & \textbf{14.00} & \textbf{40.57} & \textbf{23.21} & \textbf{40.6} & \textbf{36.37} & \textbf{0.0436}\\ \hline
\multirow{3}{*}{Car}  & pSp  & 0.115          & 0.298          & 17.24          & 19.76          & 27.25          & 36.01          & 238.0         & 66.11          & 0.0565\\
                      & e4e  & 0.110          & 0.314          & 14.68          & 18.25          & 21.50          & 27.57          & 238.0         & 66.11          & 0.0541\\
                      & Ours & \textbf{0.089} & \textbf{0.245} & \textbf{13.58} & \textbf{16.14} & \textbf{21.24} & \textbf{25.28} & \textbf{40.6} & \textbf{36.34} & \textbf{0.0435}\\ \hline
\end{tabular}
}
\vspace{-0.1cm}
\caption{Quantitative comparison for different inversion methods. To consider the distortion-editability trade-off, we list metrics for image editing to give a comprehensive evaluation on them. We also list the parameters and FLOPs of the three methods, \textit{Time} means the inference time of an iteration.}
\vspace{-0.4cm}
\label{tab:inversion}
\end{table*}

\begin{figure*}[ht]
\centering
\includegraphics[width=.9\textwidth]{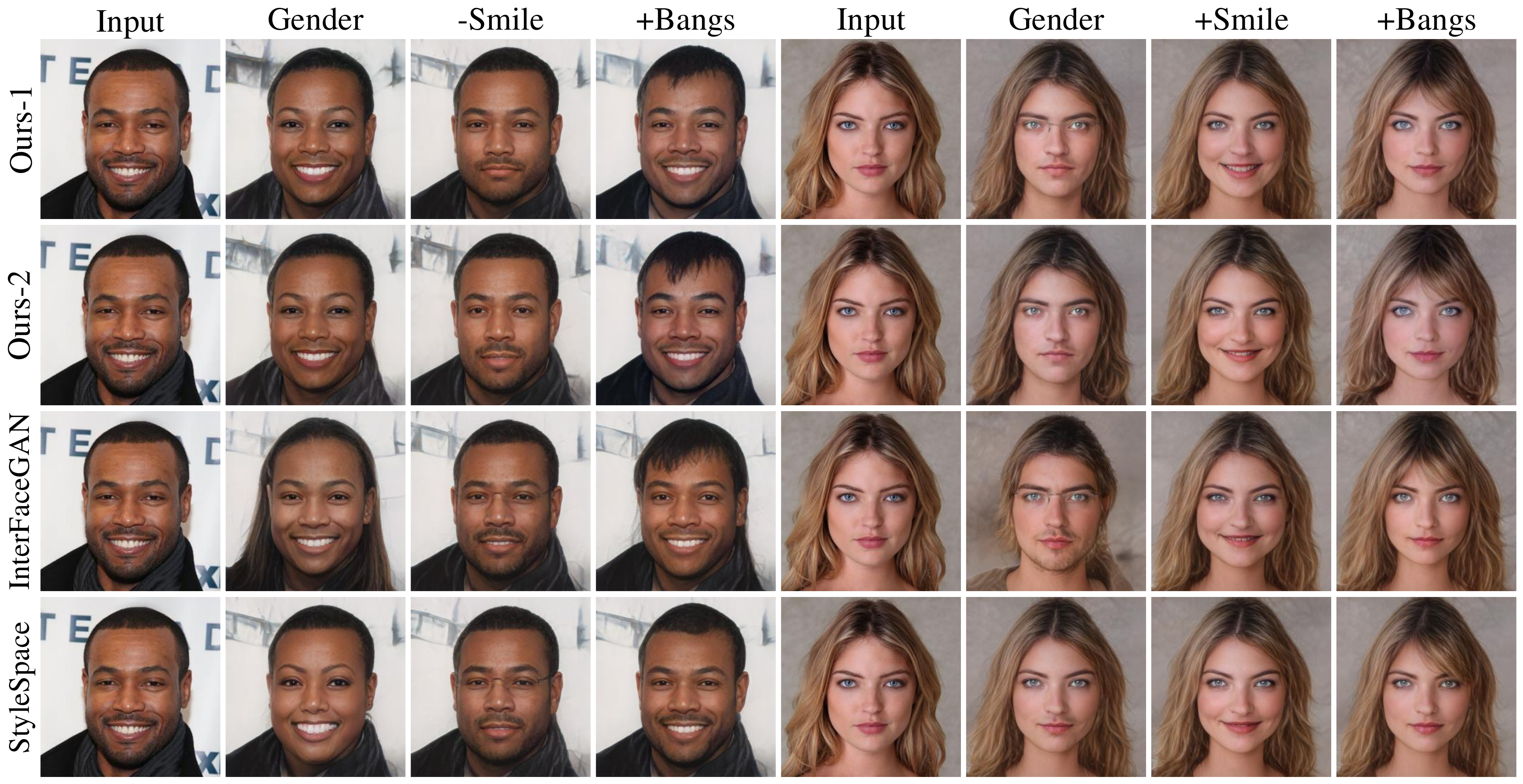}
\vspace{-0.2cm}
\caption{Qualitative comparison on different label-based editing methods. We list the results for editing "Gender", "Smile" and "Bangs", and compare them with \cite{shen2020interfacegan,wu2021stylespace}. Note that we evaluate both the first- and second-order editing methods proposed in \cref{sec:labelbased}.}
\label{fig:labelresults}
\vspace{-0.2cm}
\end{figure*}

\section{Experiments}
\subsection{Implementation Details}
All experiments are implemented on StyleGAN2 \cite{karras2020analyzing} pretrained on FFHQ \cite{karras2019style} and LSUN Cars \cite{yu2015lsun} datasets. We build our model based on the pSp encoder for multi-scale image feature. For face domain, we train the inversion model on FFHQ dataset and evaluate on CelebA-HQ \cite{karras2017progressive} test set. For car domain, the inversion model is trained and evaluated on Stanford Cars \cite{krause20133d} dataset. The synthesis network in StyleGAN2 is fixed and all other parameters in our model is trainable.

\subsection{Inversion Results}
We compare our model with pSp \cite{richardson2021encoding} and e4e \cite{tov2021designing}, which are two state-of-the-art encoder-based inversion methods. Qualitative and quantitative results are shown in \cref{fig:inversion} and \cref{tab:inversion}. Our model is validated in three aspects: the perceptual quality of inversion, the ability of editing, and the model size. MSE and LPIPS evaluate the pixel and perceptual similarity of input and inverted images. FID \cite{heusel2017gans} and SWD \cite{rabin2011wasserstein} measure the distance between two distributions of real and generated images, indicating the visual quality of generated images. To compare the editing ability of three methods, we adopt InterFaceGAN \cite{shen2020interfacegan} in face domain to edit the latent codes generated by each method. For car domain, we apply GANSpace \cite{harkonen2020ganspace} to find the semantic directions. The metrics are averaged over the editing results of the whole test set. 
%five attributes: Smile, Bangs, Eyeglasses, Gender and Makeup. 
Our model is outperformed in inversion and of higher editability. Moreover, we list the parameters, FLOPs and inference time of three methods in \cref{tab:inversion}. Compared with Convnet, the transformer used in our model has only 18 or 16 tokens for face and car domain, hence it is lightweight and efficient.

\begin{table}[t]
\centering
\resizebox{.9\columnwidth}{!}{
\begin{tabular}{p{18mm}p{5.5mm}<{\centering}p{6.5mm}<{\centering}p{5.5mm}<{\centering}p{6.5mm}<{\centering}p{5.5mm}<{\centering}p{6.5mm}<{\centering}}
% \begin{tabular}{lcccccc}
\toprule
\multirow{2}{*}{\textbf{Method}} & \multicolumn{2}{c}{\textbf{Gender}} & \multicolumn{2}{c}{\textbf{Smile}} & \multicolumn{2}{c}{\textbf{Bangs}} \\ \cmidrule(l){2-7} 
                                 & FID$\downarrow$ & SWD$\downarrow$ & FID$\downarrow$ & SWD$\downarrow$ & FID$\downarrow$ & SWD$\downarrow$ \\ \midrule
InterFaceGAN                     & 48.72          & 19.43          & 40.03          & 18.94          & 44.01          & 29.41          \\
StyleSpace                       & 37.31          & 17.31          & 34.72          & 15.46          & 42.91          & 20.96          \\
Ours-1                           & 38.73          & 17.83          & 33.50          & \textbf{14.89} & 41.15          & 19.30          \\
Ours-2                           & \textbf{34.84} & \textbf{16.14} & \textbf{32.88} & 15.23          & \textbf{40.14} & \textbf{18.53} \\ \bottomrule
\end{tabular}
}
\vspace{-0.2cm}
\caption{Quantitative comparison of label-based editing on three attributes,  corresponding to \cref{fig:labelresults}. }
\label{tab:labelfid}
\vspace{-0.1cm}
\end{table}

\begin{figure}[ht]
\centering
\includegraphics[width=1\columnwidth]{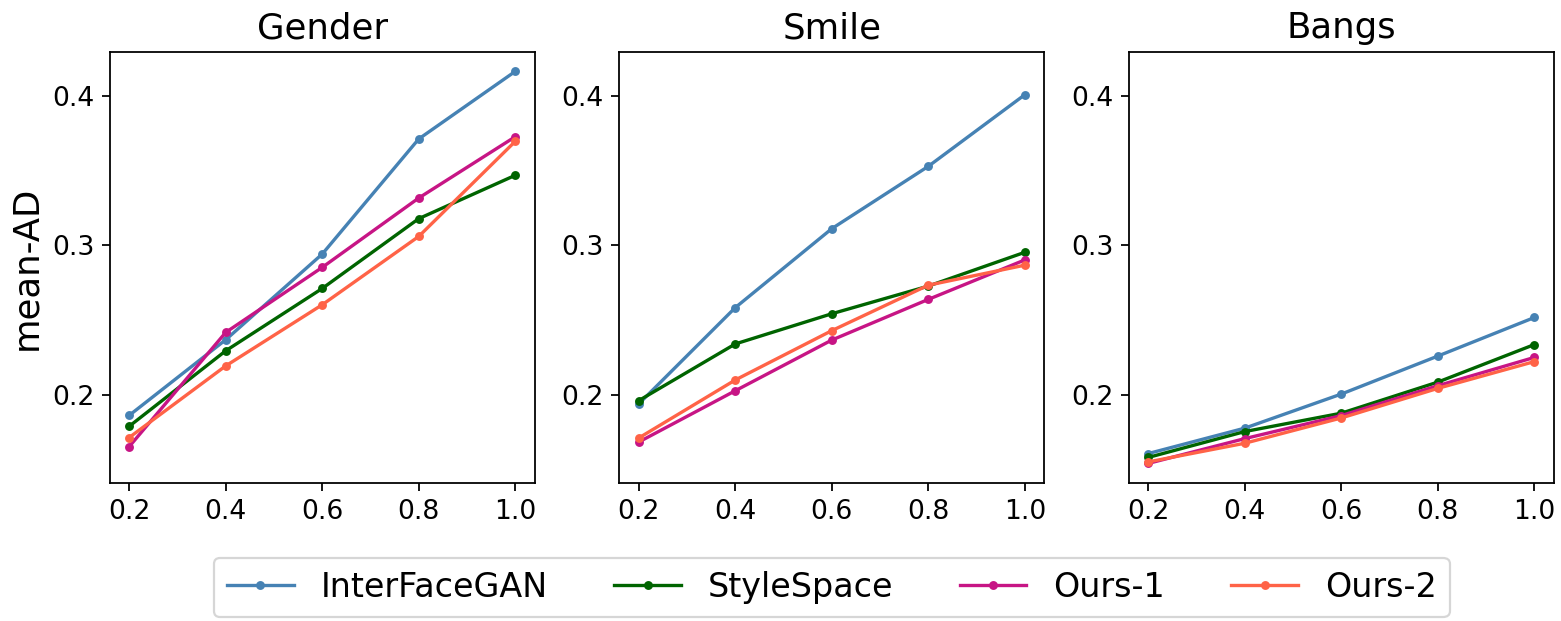}
\vspace{-0.3cm}
\caption{Mean-AD results of label-based editing on three attributes compared with \cite{shen2020interfacegan,wu2021stylespace}, lower means better. Ours-1 and Ours-2 represent our first- and second-order methods, respectively.}
\label{fig:meanad}
\vspace{-0.3cm}
\end{figure}

\begin{figure*}[!htbp]
\centering
\includegraphics[width=.9\textwidth]{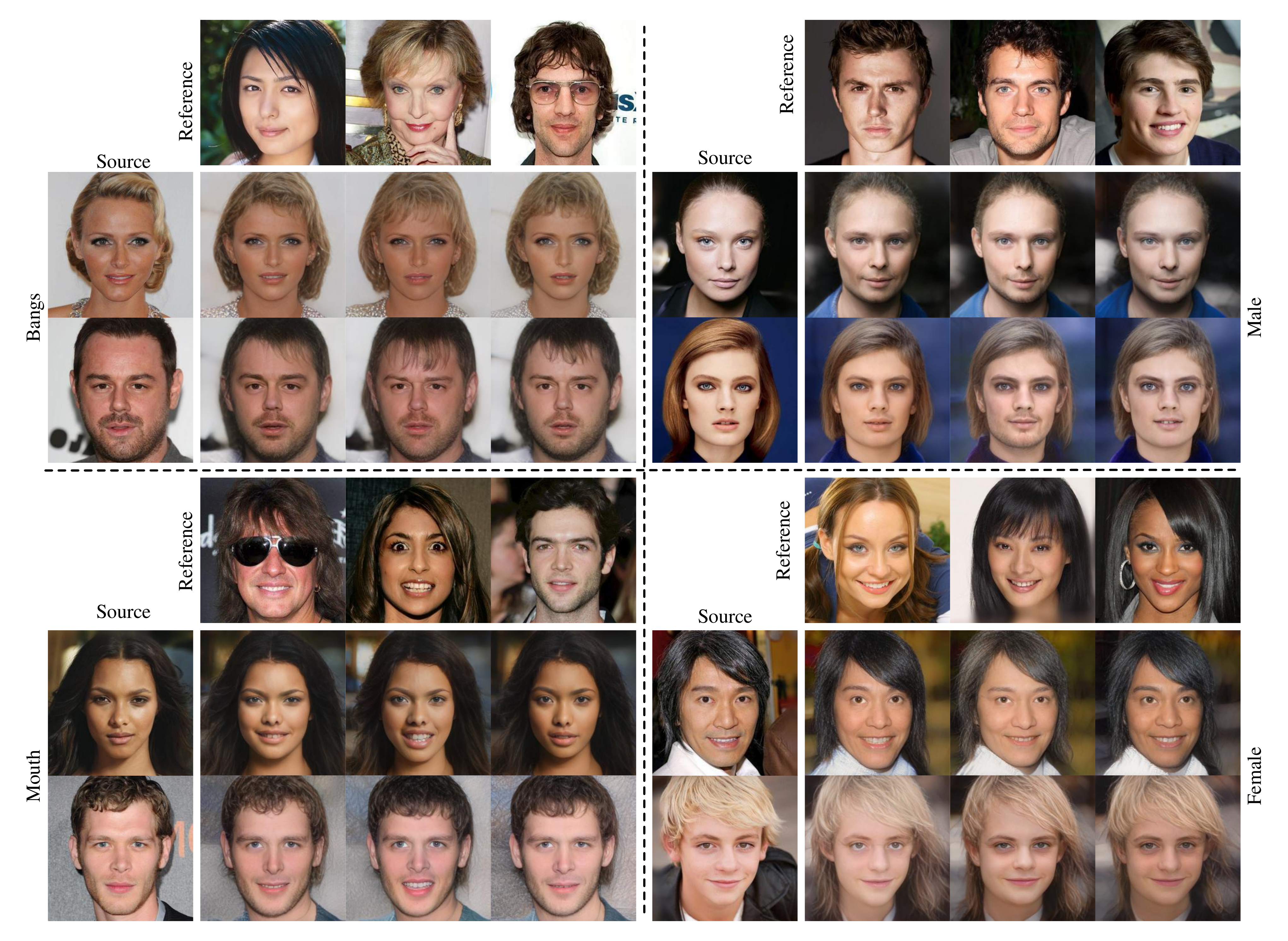}
\vspace{-0.1cm}
\caption{Reference-based editing results. Given a pair of source and reference images, we first utilize the proposed method to find their inverted codes in $W^+$. Then the transformer block described in \cref{sec:refbased} is used to take the "bangs", "mouth" and "gender" style in the reference code, and apply them onto the source.}
\label{fig:refresults}
\vspace{-0.3cm}
\end{figure*}

\subsection{Editing Results}
We apply reference- and label-based editing on CelebA-HQ dataset, in which each image has the label of 40 facial attributes. We invert images to latent codes using our pretrained inversion model, and train a 40-class latent classifier. The latent classifier consists of 4 fully-connected layers, in which there is an independent branch for each attribute before the prediction, leading to the independent embedding feature. 

\indent\textbf{Label-based Editing.}
We first apply our pretrained inversion model to obtain the latent codes of images, and use the first- and second-order methods to edit the images to have the target attributes. Desirable results can be generated by only ONE iteration. We evaluate first- and second-order methods illustrated in \cref{sec:labelbased} and compare our results with InterFaceGAN \cite{shen2020interfacegan} and StyleSpace \cite{wu2021stylespace}. Qualitative results and metrics are shown in \cref{fig:labelresults} and \cref{tab:labelfid}. Moreover, we measure the disentanglement of attributes by calculating the Attribute Dependency (AD) \cite{wu2021stylespace}, which indicates the degree of changes in other attributes while editing one attribute. A multi-branch attribute classifier based on ResNet-50 \cite{he2016deep} is applied to obtain the predicted logits of images. We measure the changes $\Delta l$ between the input and edited images, and normalize $\Delta l$ by $\sigma(l)$, which is the standard deviation computed from the logits of numerous generated images. For a target attribute $k$, we calculate the mean-AD on other attributes $\cancel{k}$ as $\mathbb{E}(\Delta l_i/\sigma(l_i))$, where $i\in \cancel{k}$ is the index of fixed attributes. \cref{fig:meanad} illustrates the mean-AD over the degree of target attribute changes $\Delta l_k/\sigma(l_k)$, Our methods perform better than InterFaceGAN and StyleSpace, and the second-order method is of higher disentanglement. 

\begin{table}[t]
\centering
\resizebox{1\columnwidth}{!}{%
\begin{tabular}{@{}lccc|ccc@{}}
\toprule
\multirow{2}{*}{\textbf{Method}} & \multicolumn{3}{c}{\textbf{Quality(\%)}}         & \multicolumn{3}{c}{\textbf{Disentanglement(\%)}} \\ \cmidrule(l){2-7} 
 &
  \multicolumn{1}{c}{\textbf{BA}} &
  \multicolumn{1}{c}{\textbf{GE}} &
  \multicolumn{1}{c}{\textbf{GO}} &
  \multicolumn{1}{c}{\textbf{BA}} &
  \multicolumn{1}{c}{\textbf{GE}} &
  \multicolumn{1}{c}{\textbf{GO}} \\ \midrule
InterFaceGAN                     & 15.00          & 7.50           & 9.17           & 11.67          & 1.67           & 8.33          \\
StyleSpace                       & 10.83          & 10.00          & 13.33          & 18.33          & 15.00          & 10.83          \\
Ours-1                           & 25.83          & 39.17          & 31.67          & \textbf{35.83} & 34.17          & 30.00          \\
Ours-2                           & \textbf{48.33} & \textbf{43.33} & \textbf{47.50} & 34.17          & \textbf{49.17} & \textbf{49.17} \\ \bottomrule
\end{tabular}%
}
\caption{User study of label-based editing compared with \cite{shen2020interfacegan}, \cite{wu2021stylespace}. \textbf{BA}, \textbf{GE} and \textbf{GO} represent ‘Bangs’, ‘Gender’ and ‘Goatee’ attributes.}
\label{tab:user}
\vspace{-0.2cm}
\end{table}

Considering human judgements, we further conduct a user study. We ask 60 volunteers to evaluate the methods in two aspects: image quality and disentanglement. Results are shown in \cref{tab:user}. The detailed algorithms of first- and second-order methods are provided in Appendix C.

\indent\textbf{Reference-based Editing.}
The reference-based editing module is trained for different attributes individually. To ensure the module takes the style from reference image, we randomly divide the training images into source and reference sets, instead of depending on the labels. We train the module on three attributes and qualitative results are shown in \cref{fig:refresults}. The edited images take the relevant attributes from different reference images, and they appear the similar style on the translated attribute. Note that the reference-based editing module is trained only in the latent space, resulting in less diversity compared with directly editing on images. Whereas, different from the optimized-based method \cite{collins2020editing}, our model can commonly apply to all images, which is lightweight and more flexible.

\begin{table}[t]
\centering
\resizebox{1\columnwidth}{!}{%
\begin{tabular}{p{19mm}<{\centering}p{7.5mm}<{\centering}p{8.5mm}<{\centering}p{15mm}<{\centering}p{13.5mm}<{\centering}p{10.5mm}<{\centering}}
\toprule
\textbf{Method} & \textbf{MSE$\downarrow$} & \textbf{LPIPS$\downarrow$} & \textbf{Params(M)$\downarrow$} & \textbf{FLOPs(G)$\downarrow$} & \textbf{Time(s)$\downarrow$} \\ \midrule
pSp           & 0.0373          & 0.1693          & 267.3         & 72.55          & 0.0668          \\
% ReStyle       & \textbf{0.0314} & \textbf{0.1379} & 205.1         & 183.12          & 0.2732          \\
Ours w/o self & 0.0369          & 0.1716          & \textbf{37.3} & \textbf{36.31} & \textbf{0.0429} \\
Ours full     & \textbf{0.0363} & \textbf{0.1665} & 40.6          & 36.37          & 0.0436          \\\bottomrule
\end{tabular}%
}
% \vspace{-0.1cm}
\caption{Ablations of transformer structure. \textit{Time} means the inference time of an iteration. The best results are indicated in \textbf{Bold}.}
\label{tab:ablation}
% \vspace{-0.2cm}
\end{table}

\subsection{Ablations and Analysis}
We further validate the benefit of transformer by comparing among pSp \cite{richardson2021encoding}, our full model with both self- and cross-attention and ours w/o self-attention in \cref{tab:ablation}. \cite{richardson2021encoding} maps image features to $w+$ by individual mapping networks, though $w+$ obtain the image features directly and completely, the relation between each $w$ is not tightly enough. In our model, cross-attention is necessary to update queries by fusing image features, and self-attention is also important in constructing the potential relation between queries.

\section{Limitations}
%Our work has some limitations. 
We now discuss limitations, which we have already realized, for our work. First, for the inversion task, although our proposed method achieves improved reconstruction quality, there are still some differences between the input %images 
and reconstructed images, especially for %when the images are 
the out-of-domain input. We think it is mainly caused by the finite discriminative ability of %the code 
$W^+$ space. As is described in \cite{wang2021HFGI}, the distortion can be significantly reduced by adding more information from the source. Moreover, since we apply the multi-head attention, the training speed is slower due to the complex matrix multiplication.
Second, for the reference-based editing task, we adopt %reference-based 
a transformer-based module in the latent space, resulting in less diversity for some attributes compared with direct editing on the images, in which the mode seeking loss \cite{mao2019mode} can encourage the diversity in the pixel domain. But our method is lightweight and more flexible.

\section{Conclusion}
This paper presents a transformer-based image inversion and editing method for StyleGAN. We choose $W^+$ space to represent real images, which needs to determine multiple style codes for different layers of the generator. To effectively exploit information from input image, we design a multi-stage transformer module, which mainly consists of the self- and cross-attention. In the initial stage, the MLP maps a set of learnable noise vectors into the codes in $W^+$, and then they are iteratively updated by the two types of attention operations, so the codes from the final stage can reconstruct the input accurately. Based on them, we are able to carry out label- and reference-based editing in a flexible way. Given a required label, an encoder-free strategy is employed to find the unique editing vector according to the gradient from a pretrained latent classifier. Meanwhile, given a reference code, a transformer block is trained to edit the source, so that the result takes the relevant style from the reference. Experiments show the proposed image inversion and editing method achieves less distortions and higher quality at the same time.

%%%%%%%%% REFERENCES
\clearpage
\appendix

{\LARGE\noindent\textbf{Appendix}}
\vspace{0.5cm}
\section{Training Details}
We adopt a pretrained StyleGAN2 \cite{karras2020analyzing} generator in our experiments, in which the synthesis network is fixed and the mapping network (MLP) is trained. In the multi-head attention of the transformer block, the number of heads is set to 4, and the dimension of each head is 512. For inversion task, the Ranger optimizer is used in training, which is a combination of Rectified Adam \cite{liu2019variance} with the Lookahead technique \cite{zhang2019lookahead}. We train the model for $6 \times 10^{5}$ iterations with a batch size of 8, the learning rate is set to $1 \times 10^{-4}$. For the reference-based editing task, we use the Adam \cite{kingma2014adam} optimizer to train the model for $1 \times 10^{4}$ iterations with a batch size of 8, the learning rate is set to $1 \times 10^{-3}$. All experiments are implemented on 2 NVIDIA RTX 2080Ti GPUs.

\section{Label-based Editing Methods}
\begin{figure}[hb]
\centering
\includegraphics[width=0.8\columnwidth]{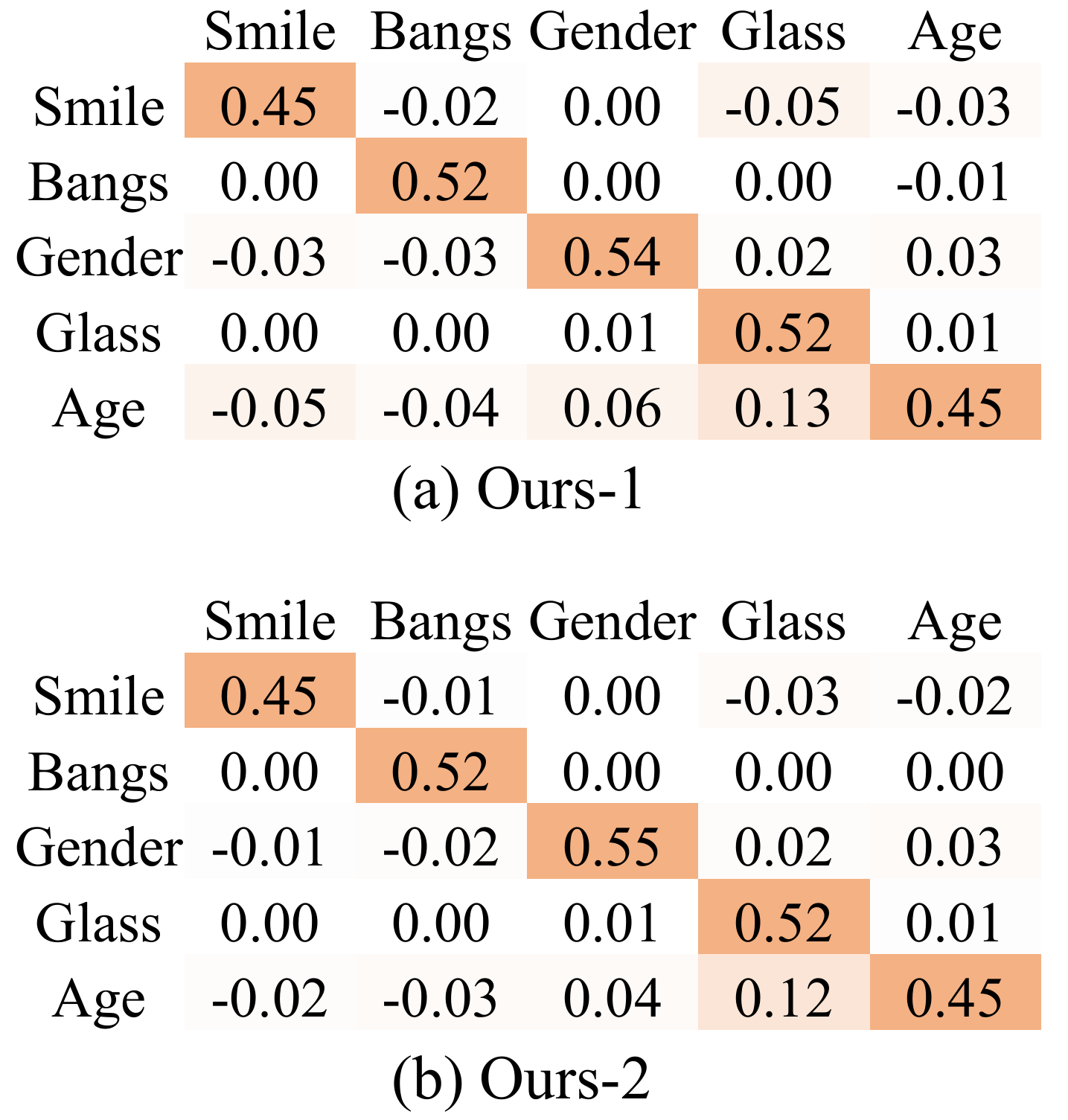}
\caption{Re-scoring results of label-based editing on five attributes, Ours-1 and Ours-2 represent our first- and second-order methods, respectively.}
\label{fig:rescoring}
\end{figure}

We propose first- and second-order label-based editing methods in the main text. To give a detailed explanation, we provide the pseudo codes in PyTorch style. \cref{alg:label1} and \cref{alg:label2} illustrate the first- and second-order methods, respectively. Moreover, we measure the disentanglement of five attributes by Re-scoring \cite{shen2020interfacegan} in \cref{fig:rescoring}. The top row lists edited attributes, and the scores are the classification logits changes between original and edited images. 

\section{More Results}
\begin{figure*}[ht]
\centering
\includegraphics[width=.9\textwidth]{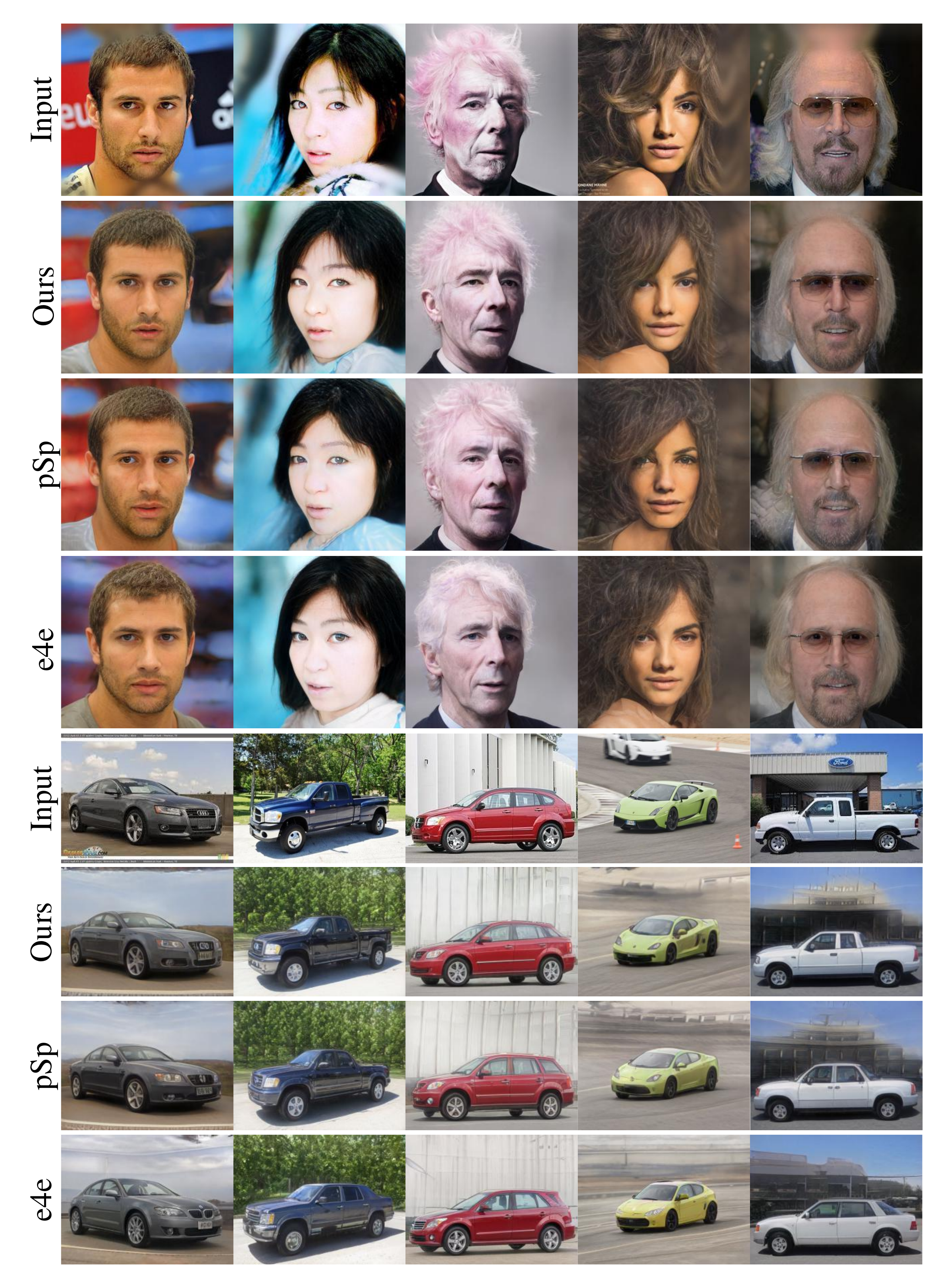}
\caption{More results of inversion compared with \cite{richardson2021encoding} and \cite{tov2021designing}.}
\label{fig:appinvert}
\end{figure*}

\begin{figure*}[ht]
\centering
\includegraphics[width=.9\textwidth]{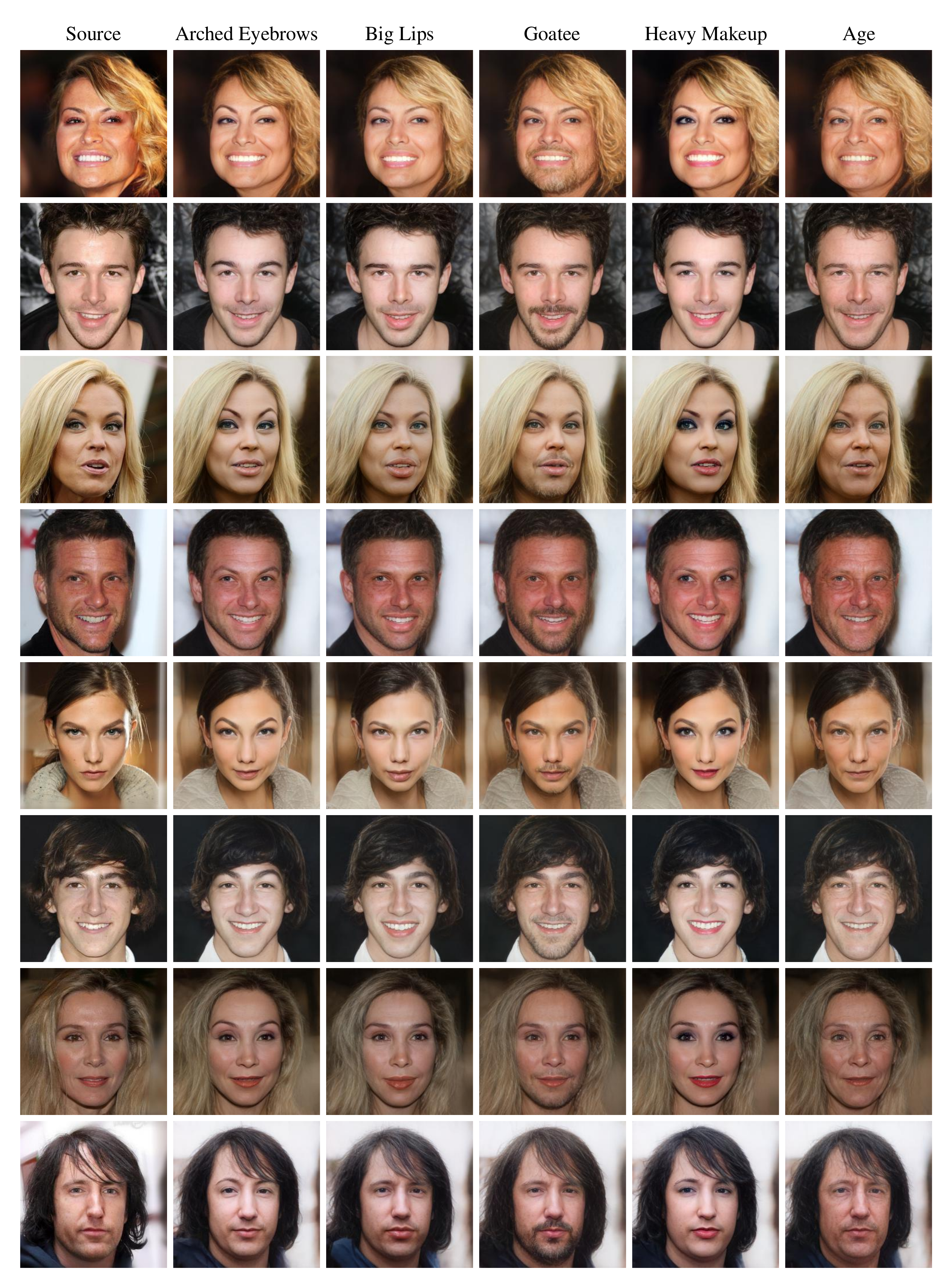}
\caption{More results of label-based editing on five attributes.}
\label{fig:applabel}
\end{figure*}

\begin{figure*}[ht]
\centering
\includegraphics[width=1\textwidth]{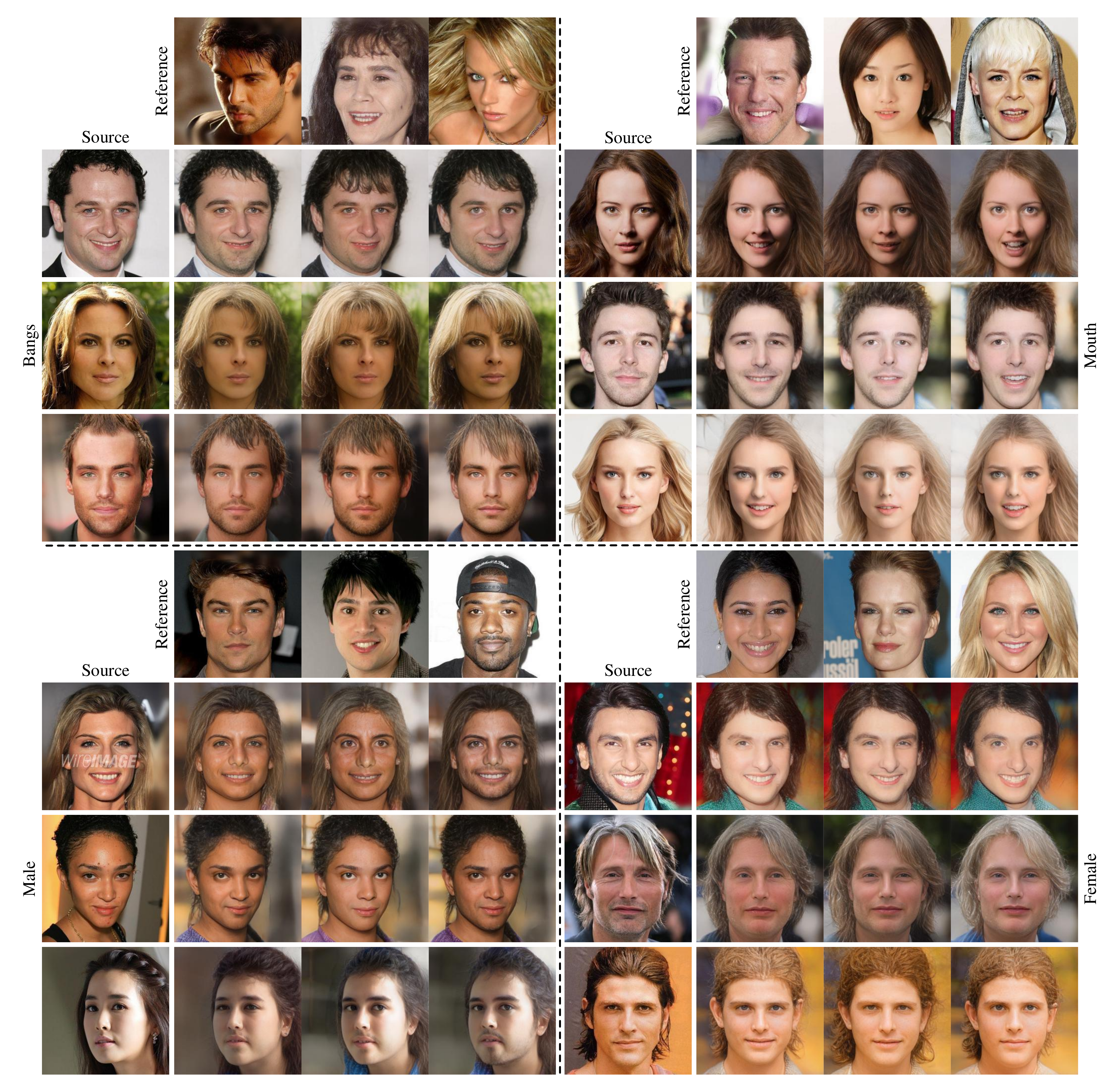}
\caption{More results of reference-based editing on three attributes. The edited images take the style of \textit{Bangs}, \textit{Mouth} and \textit{Gender} from different reference images.}
\label{fig:appref}
\end{figure*}

In this section, we provide more results of inversion, label-based editing and reference-based editing in \cref{fig:appinvert}, \cref{fig:applabel}, \cref{fig:appref}.

\vspace{0.2cm}
\begin{algorithm}[ht]
\caption{{First-order Label-based Editing}}
\label{alg:label1}

\begin{lstlisting}[language=python]
# w: input latent code (18, 512)
# C: latent classifier
# y_t: target label

predicted = C(w)
loss = torch.nn.BCELoss(predicted, y_t)
loss.backward()
direct = w.grad
direct = direct / torch.norm(direct, dim=1)
w_edit = w - alpha * direct # alpha is a scaling factor.
\end{lstlisting}
\end{algorithm}

\begin{algorithm}[ht]
\caption{{Second-order Label-based Editing}}
\label{alg:label2}
\begin{lstlisting}[language=python]
# w: input latent code (18, 512)
# C: latent classifier
# y_t: target label

r_d = torch.randn(18, 512)
r_0 = torch.zeros(18, 512)
w_d = w + kasi * r_d # kasi is a small number, we set it to 10e-4.
w_0 = w + r_0
predicted_d = C(w_d)
loss = torch.nn.BCELoss(predicted_d, y_t)
loss.backward()
direct_d = r_d.grad

C.zero_grad()
predicted_0 = C(w_0)
loss = torch.nn.BCELoss(predicted_0, y_t)
loss.backward
direct_0 = r_0.grad

direct = direct_d - direct_0
direct = direct / torch.norm(direct, dim=1)
w_edit = w - alpha * direct # alpha is a scaling factor.
\end{lstlisting}
% \vspace{-0.2cm}
\end{algorithm}
% \vspace{-0.2cm}

{\small
\begin{thebibliography}{10}\itemsep=-1pt
	
	\bibitem{abdal2019image2stylegan}
	Rameen Abdal, Yipeng Qin, and Peter Wonka.
	\newblock Image2stylegan: How to embed images into the stylegan latent space?
	\newblock In {\em Proceedings of the IEEE/CVF International Conference on
		Computer Vision}, pages 4432--4441, 2019.
	
	\bibitem{abdal2020image2stylegan++}
	Rameen Abdal, Yipeng Qin, and Peter Wonka.
	\newblock Image2stylegan++: How to edit the embedded images?
	\newblock In {\em Proceedings of the IEEE/CVF Conference on Computer Vision and
		Pattern Recognition}, pages 8296--8305, 2020.
	
	\bibitem{abdal2021styleflow}
	Rameen Abdal, Peihao Zhu, Niloy~J Mitra, and Peter Wonka.
	\newblock Styleflow: Attribute-conditioned exploration of stylegan-generated
	images using conditional continuous normalizing flows.
	\newblock {\em ACM Transactions on Graphics (TOG)}, 40(3):1--21, 2021.
	
	\bibitem{alaluf2021restyle}
	Yuval Alaluf, Or Patashnik, and Daniel Cohen-Or.
	\newblock Restyle: A residual-based stylegan encoder via iterative refinement.
	\newblock In {\em Proceedings of the IEEE/CVF International Conference on
		Computer Vision}, pages 6711--6720, 2021.
	
	\bibitem{brock2018large}
	Andrew Brock, Jeff Donahue, and Karen Simonyan.
	\newblock Large scale gan training for high fidelity natural image synthesis.
	\newblock {\em arXiv preprint arXiv:1809.11096}, 2018.
	
	\bibitem{carion2020end}
	Nicolas Carion, Francisco Massa, Gabriel Synnaeve, Nicolas Usunier, Alexander
	Kirillov, and Sergey Zagoruyko.
	\newblock End-to-end object detection with transformers.
	\newblock In {\em European Conference on Computer Vision}, pages 213--229.
	Springer, 2020.
	
	\bibitem{choi2020stargan}
	Yunjey Choi, Youngjung Uh, Jaejun Yoo, and Jung-Woo Ha.
	\newblock Stargan v2: Diverse image synthesis for multiple domains.
	\newblock In {\em Proceedings of the IEEE/CVF Conference on Computer Vision and
		Pattern Recognition}, pages 8188--8197, 2020.
	
	\bibitem{collins2020editing}
	Edo Collins, Raja Bala, Bob Price, and Sabine Susstrunk.
	\newblock Editing in style: Uncovering the local semantics of gans.
	\newblock In {\em Proceedings of the IEEE/CVF Conference on Computer Vision and
		Pattern Recognition}, pages 5771--5780, 2020.
	
	\bibitem{deng2019arcface}
	Jiankang Deng, Jia Guo, Niannan Xue, and Stefanos Zafeiriou.
	\newblock Arcface: Additive angular margin loss for deep face recognition.
	\newblock In {\em Proceedings of the IEEE/CVF Conference on Computer Vision and
		Pattern Recognition}, pages 4690--4699, 2019.
	
	\bibitem{dosovitskiy2020image}
	Alexey Dosovitskiy, Lucas Beyer, Alexander Kolesnikov, Dirk Weissenborn,
	Xiaohua Zhai, Thomas Unterthiner, Mostafa Dehghani, Matthias Minderer, Georg
	Heigold, Sylvain Gelly, et~al.
	\newblock An image is worth 16x16 words: Transformers for image recognition at
	scale.
	\newblock {\em arXiv preprint arXiv:2010.11929}, 2020.
	
	\bibitem{harkonen2020ganspace}
	Erik H{\"a}rk{\"o}nen, Aaron Hertzmann, Jaakko Lehtinen, and Sylvain Paris.
	\newblock Ganspace: Discovering interpretable gan controls.
	\newblock {\em arXiv preprint arXiv:2004.02546}, 2020.
	
	\bibitem{he2016deep}
	Kaiming He, Xiangyu Zhang, Shaoqing Ren, and Jian Sun.
	\newblock Deep residual learning for image recognition.
	\newblock In {\em Proceedings of the IEEE conference on computer vision and
		pattern recognition}, pages 770--778, 2016.
	
	\bibitem{heusel2017gans}
	Martin Heusel, Hubert Ramsauer, Thomas Unterthiner, Bernhard Nessler, and Sepp
	Hochreiter.
	\newblock Gans trained by a two time-scale update rule converge to a local nash
	equilibrium.
	\newblock {\em Advances in neural information processing systems}, 30, 2017.
	
	\bibitem{huang2021bridging}
	Qiusheng Huang, Zhilin Zheng, Xueqi Hu, Li Sun, and Qingli Li.
	\newblock Bridging the gap between label-and reference-based synthesis in
	multi-attribute image-to-image translation.
	\newblock In {\em Proceedings of the IEEE/CVF International Conference on
		Computer Vision}, pages 14628--14637, 2021.
	
	\bibitem{huang2017arbitrary}
	Xun Huang and Serge Belongie.
	\newblock Arbitrary style transfer in real-time with adaptive instance
	normalization.
	\newblock In {\em Proceedings of the IEEE International Conference on Computer
		Vision}, pages 1501--1510, 2017.
	
	\bibitem{karras2017progressive}
	Tero Karras, Timo Aila, Samuli Laine, and Jaakko Lehtinen.
	\newblock Progressive growing of gans for improved quality, stability, and
	variation.
	\newblock {\em arXiv preprint arXiv:1710.10196}, 2017.
	
	\bibitem{karras2019style}
	Tero Karras, Samuli Laine, and Timo Aila.
	\newblock A style-based generator architecture for generative adversarial
	networks.
	\newblock In {\em Proceedings of the IEEE/CVF Conference on Computer Vision and
		Pattern Recognition}, pages 4401--4410, 2019.
	
	\bibitem{karras2020analyzing}
	Tero Karras, Samuli Laine, Miika Aittala, Janne Hellsten, Jaakko Lehtinen, and
	Timo Aila.
	\newblock Analyzing and improving the image quality of stylegan.
	\newblock In {\em Proceedings of the IEEE/CVF Conference on Computer Vision and
		Pattern Recognition}, pages 8110--8119, 2020.
	
	\bibitem{kim2021exploiting}
	Hyunsu Kim, Yunjey Choi, Junho Kim, Sungjoo Yoo, and Youngjung Uh.
	\newblock Exploiting spatial dimensions of latent in gan for real-time image
	editing.
	\newblock In {\em Proceedings of the IEEE/CVF Conference on Computer Vision and
		Pattern Recognition}, pages 852--861, 2021.
	
	\bibitem{kingma2014adam}
	Diederik~P Kingma and Jimmy Ba.
	\newblock Adam: A method for stochastic optimization.
	\newblock {\em arXiv preprint arXiv:1412.6980}, 2014.
	
	\bibitem{krause20133d}
	Jonathan Krause, Michael Stark, Jia Deng, and Li Fei-Fei.
	\newblock 3d object representations for fine-grained categorization.
	\newblock In {\em Proceedings of the IEEE international conference on computer
		vision workshops}, pages 554--561, 2013.
	
	\bibitem{liu2019variance}
	Liyuan Liu, Haoming Jiang, Pengcheng He, Weizhu Chen, Xiaodong Liu, Jianfeng
	Gao, and Jiawei Han.
	\newblock On the variance of the adaptive learning rate and beyond.
	\newblock {\em arXiv preprint arXiv:1908.03265}, 2019.
	
	\bibitem{liu2021swin}
	Ze Liu, Yutong Lin, Yue Cao, Han Hu, Yixuan Wei, Zheng Zhang, Stephen Lin, and
	Baining Guo.
	\newblock Swin transformer: Hierarchical vision transformer using shifted
	windows.
	\newblock {\em arXiv preprint arXiv:2103.14030}, 2021.
	
	\bibitem{locatello2020object}
	Francesco Locatello, Dirk Weissenborn, Thomas Unterthiner, Aravindh Mahendran,
	Georg Heigold, Jakob Uszkoreit, Alexey Dosovitskiy, and Thomas Kipf.
	\newblock Object-centric learning with slot attention.
	\newblock {\em arXiv preprint arXiv:2006.15055}, 2020.
	
	\bibitem{mao2019mode}
	Qi Mao, Hsin-Ying Lee, Hung-Yu Tseng, Siwei Ma, and Ming-Hsuan Yang.
	\newblock Mode seeking generative adversarial networks for diverse image
	synthesis.
	\newblock In {\em Proceedings of the IEEE/CVF Conference on Computer Vision and
		Pattern Recognition}, pages 1429--1437, 2019.
	
	\bibitem{miyato2018virtual}
	Takeru Miyato, Shin-ichi Maeda, Masanori Koyama, and Shin Ishii.
	\newblock Virtual adversarial training: a regularization method for supervised
	and semi-supervised learning.
	\newblock {\em IEEE transactions on pattern analysis and machine intelligence},
	41(8):1979--1993, 2018.
	
	\bibitem{oord2017neural}
	Aaron van~den Oord, Oriol Vinyals, and Koray Kavukcuoglu.
	\newblock Neural discrete representation learning.
	\newblock {\em arXiv preprint arXiv:1711.00937}, 2017.
	
	\bibitem{rabin2011wasserstein}
	Julien Rabin, Gabriel Peyr{\'e}, Julie Delon, and Marc Bernot.
	\newblock Wasserstein barycenter and its application to texture mixing.
	\newblock In {\em International Conference on Scale Space and Variational
		Methods in Computer Vision}, pages 435--446. Springer, 2011.
	
	\bibitem{razavi2019generating}
	Ali Razavi, Aaron van~den Oord, and Oriol Vinyals.
	\newblock Generating diverse high-fidelity images with vq-vae-2.
	\newblock In {\em Advances in neural information processing systems}, pages
	14866--14876, 2019.
	
	\bibitem{richardson2021encoding}
	Elad Richardson, Yuval Alaluf, Or Patashnik, Yotam Nitzan, Yaniv Azar, Stav
	Shapiro, and Daniel Cohen-Or.
	\newblock Encoding in style: a stylegan encoder for image-to-image translation.
	\newblock In {\em Proceedings of the IEEE/CVF Conference on Computer Vision and
		Pattern Recognition}, pages 2287--2296, 2021.
	
	\bibitem{shen2020interfacegan}
	Yujun Shen, Ceyuan Yang, Xiaoou Tang, and Bolei Zhou.
	\newblock Interfacegan: Interpreting the disentangled face representation
	learned by gans.
	\newblock {\em IEEE transactions on pattern analysis and machine intelligence},
	2020.
	
	\bibitem{shen2021closed}
	Yujun Shen and Bolei Zhou.
	\newblock Closed-form factorization of latent semantics in gans.
	\newblock In {\em Proceedings of the IEEE/CVF Conference on Computer Vision and
		Pattern Recognition}, pages 1532--1540, 2021.
	
	\bibitem{tov2021designing}
	Omer Tov, Yuval Alaluf, Yotam Nitzan, Or Patashnik, and Daniel Cohen-Or.
	\newblock Designing an encoder for stylegan image manipulation.
	\newblock {\em ACM Transactions on Graphics (TOG)}, 40(4):1--14, 2021.
	
	\bibitem{voynov2020unsupervised}
	Andrey Voynov and Artem Babenko.
	\newblock Unsupervised discovery of interpretable directions in the gan latent
	space.
	\newblock In {\em International Conference on Machine Learning}, pages
	9786--9796. PMLR, 2020.
	
	\bibitem{wang2021attribute}
	Rui Wang, Jian Chen, Gang Yu, Li Sun, Changqian Yu, Changxin Gao, and Nong
	Sang.
	\newblock Attribute-specific control units in stylegan for fine-grained image
	manipulation.
	\newblock In {\em Proceedings of the 29th ACM International Conference on
		Multimedia}, pages 926--934, 2021.
	
	\bibitem{wang2021HFGI}
	Tengfei Wang, Yong Zhang, Yanbo Fan, Jue Wang, and Qifeng Chen.
	\newblock High-fidelity gan inversion for image attribute editing.
	\newblock {\em arxiv:2109.06590}, 2021.
	
	\bibitem{wu2021stylespace}
	Zongze Wu, Dani Lischinski, and Eli Shechtman.
	\newblock Stylespace analysis: Disentangled controls for stylegan image
	generation.
	\newblock In {\em Proceedings of the IEEE/CVF Conference on Computer Vision and
		Pattern Recognition}, pages 12863--12872, 2021.
	
	\bibitem{yao2021latent}
	Xu Yao, Alasdair Newson, Yann Gousseau, and Pierre Hellier.
	\newblock A latent transformer for disentangled face editing in images and
	videos.
	\newblock In {\em Proceedings of the IEEE/CVF International Conference on
		Computer Vision}, pages 13789--13798, 2021.
	
	\bibitem{yu2015lsun}
	Fisher Yu, Ari Seff, Yinda Zhang, Shuran Song, Thomas Funkhouser, and Jianxiong
	Xiao.
	\newblock Lsun: Construction of a large-scale image dataset using deep learning
	with humans in the loop.
	\newblock {\em arXiv preprint arXiv:1506.03365}, 2015.
	
	\bibitem{yuksel2021latentclr}
	O{\u{g}}uz~Kaan Y{\"u}ksel, Enis Simsar, Ezgi~G{\"u}lperi Er, and Pinar
	Yanardag.
	\newblock Latentclr: A contrastive learning approach for unsupervised discovery
	of interpretable directions.
	\newblock {\em arXiv preprint arXiv:2104.00820}, 2021.
	
	\bibitem{zhang2019lookahead}
	Michael~R Zhang, James Lucas, Geoffrey Hinton, and Jimmy Ba.
	\newblock Lookahead optimizer: k steps forward, 1 step back.
	\newblock {\em arXiv preprint arXiv:1907.08610}, 2019.
	
	\bibitem{zhang2018unreasonable}
	Richard Zhang, Phillip Isola, Alexei~A Efros, Eli Shechtman, and Oliver Wang.
	\newblock The unreasonable effectiveness of deep features as a perceptual
	metric.
	\newblock In {\em Proceedings of the IEEE conference on computer vision and
		pattern recognition}, pages 586--595, 2018.
	
	\bibitem{zhu2020domain}
	Jiapeng Zhu, Yujun Shen, Deli Zhao, and Bolei Zhou.
	\newblock In-domain gan inversion for real image editing.
	\newblock In {\em European conference on computer vision}, pages 592--608.
	Springer, 2020.
	
	\bibitem{zhu2016generative}
	Jun-Yan Zhu, Philipp Kr{\"a}henb{\"u}hl, Eli Shechtman, and Alexei~A Efros.
	\newblock Generative visual manipulation on the natural image manifold.
	\newblock In {\em European conference on computer vision}, pages 597--613.
	Springer, 2016.
	
	\bibitem{zhu2020improved}
	Peihao Zhu, Rameen Abdal, Yipeng Qin, John Femiani, and Peter Wonka.
	\newblock Improved stylegan embedding: Where are the good latents?
	\newblock {\em arXiv preprint arXiv:2012.09036}, 2020.
	
	\bibitem{zhu2020deformable}
	Xizhou Zhu, Weijie Su, Lewei Lu, Bin Li, Xiaogang Wang, and Jifeng Dai.
	\newblock Deformable detr: Deformable transformers for end-to-end object
	detection.
	\newblock {\em arXiv preprint arXiv:2010.04159}, 2020.
	
\end{thebibliography}

}
\end{document}